\documentclass{article}
\usepackage{spconf,amsmath,graphicx,amssymb}

\usepackage{xcolor}
\usepackage{enumitem}
\setlist{nosep, leftmargin=14pt}
\usepackage{mwe} 
\usepackage{tabularx,booktabs}
\usepackage{multirow}
\usepackage{url}
\usepackage[usestackEOL]{stackengine}

\DeclareMathOperator*{\argmin}{arg\,min} 

\title{A Clinically Inspired Approach for Melanoma classification}
\name{Prathyusha Akundi*\thanks{*Equally contributed. Work done as part of research at CVIT, IIIT-H.}, Soumyasis Gun*, Jayanthi Sivaswamy}
\address{International Institute of Information Technology, Hyderabad\\
Center for Vision Information Technology (CVIT)\\
Hyderabad, India}
\begin{document}
\strutlongstacks{T}
%
\maketitle
\begin{abstract}
Melanoma is a leading cause of deaths due to skin cancer deaths and hence, early and effective diagnosis of melanoma is of interest. Current approaches for automated diagnosis of melanoma either use pattern recognition or analytical recognition like ABCDE (asymmetry, border, color, diameter and evolving) criterion. In practice however, a differential approach wherein outliers (ugly duckling) are detected and used to evaluate nevi/lesions.
Incorporation of differential recognition in Computer Aided Diagnosis (CAD) systems has not been explored but can be beneficial as it can provide a clinical justification for the derived decision. We present a method for identifying and quantifying ugly ducklings by performing Intra-Patient Comparative Analysis (IPCA) of neighboring nevi. This is then incorporated in a CAD system design for melanoma detection. This design ensures flexibility to handle cases where IPCA is not possible. Our experiments on a public dataset show that the outlier information helps boost the sensitivity of detection by at least 4.1\% and specificity by 4.0\% to 8.9\%, depending on the use of a strong (EfficientNet) or moderately strong (VGG and ResNet) classifier.
\end{abstract}
\section{Introduction}
Melanoma is an aggressive and lethal form of skin cancer contributing to a majority of skin cancer fatalities. Its early detection is crucial to increase the chances of survival by a large margin. Since patients can have several lesions in some specific locations 
diagnosis is a time-intensive task for melanoma lesions which can lead to misclassifications, motivating the need for a CAD.

Typically, clinicans perform three major tasks while evaluating a skin lesion: overall examination of the lesion i.e. Lesion Focussed Analysis (LFA) 
lesion-level evaluation by analytical methods such as 7-Point Checklist \cite{sevenpt}, ABCDE approach \cite{abcd_original} and an IPCA in a region, to identify any \textit{Ugly Ducklings}. In CAD development, deep learning solutions have been proposed for diagnosing melanoma by performing only LFA \cite{gru_paper}, \cite{gan_paper} as well as a combination of LFA and 7-point checklist \cite{kawahara}.


When numerous nevi (moles) are present in a region, it has been observed that melanoma lesions
deviate from the rest and exhibit very different properties. Hence, detecting outliers or \emph{Ugly Ducklings} has been argued \cite{uglyduckling} \cite{GaudyMarqueste2017UglyDS}, to improve the effectiveness of screening and early detection. Public awareness campaigns about melanoma also advise individuals with nevi to perform a self-exam to identify the ugly ducklings and report the same to medical experts.

Since a differential recognition has not been explored in CAD development for melanoma detection, there are 2 questions of interest. First, is on whether outlier detection can be beneficial to melanoma classification by a CAD system. It is  possible that either multiple nevi are not present in a patient or a CAD has no access to patient-level information to identify contextual lesions. This leads to the second question, namely on how to design a CAD system that is flexible in handling a variable input, namely, a singleton set, or a contextual set of lesions. 
In this paper, we answer these questions with the following contributions. 
\begin{itemize}
\item A solution for outlier detection based on a pretrained CNN. The output of this module is an outlier score for each of the contextual lesions.  Such a score can aid clinicians to decide which of the lesions are to be excised and help reduce the excision rate of benign lesions while adding explanation for the inference. 
\item A modular design for a CAD system for melanoma detection. The design includes a solution for the integration of outlier detection module with a module for melanoma classification which can be a pre-existing system or a new design. This integration permits the functioning of the overall system \textit{even in the absence} of contextual lesions information.
\end{itemize}



\begin{figure*}[htb]

\begin{minipage}[b]{.48\linewidth}
  \centering
  \centerline{\includegraphics[width=10.0cm]{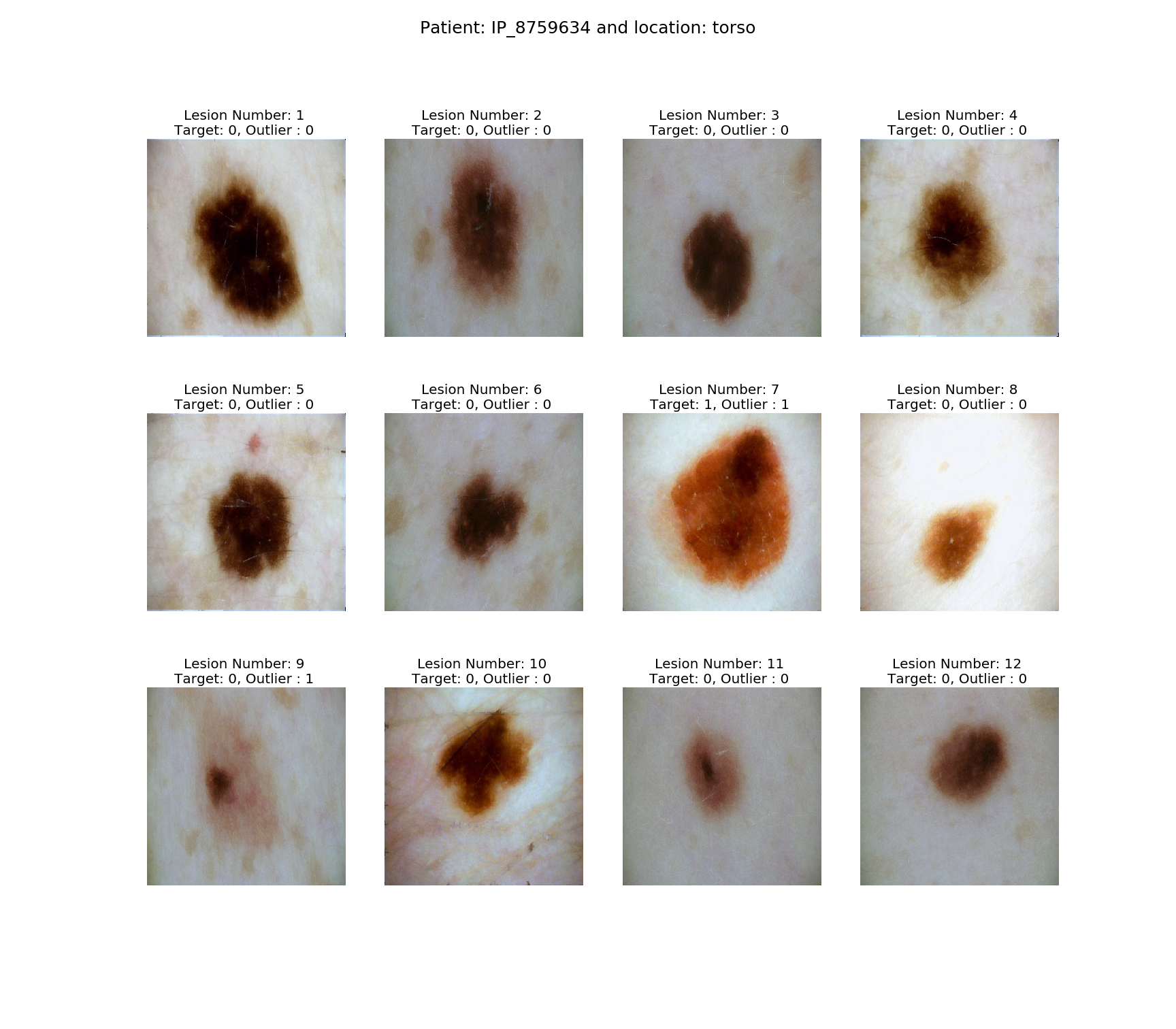}}
  \centerline{(a) Contextual Lesions from the torso of a patient}\medskip
\end{minipage}
\hfill
\begin{minipage}[b]{0.48\linewidth}
  \centering
  \centerline{\includegraphics[width=10cm]{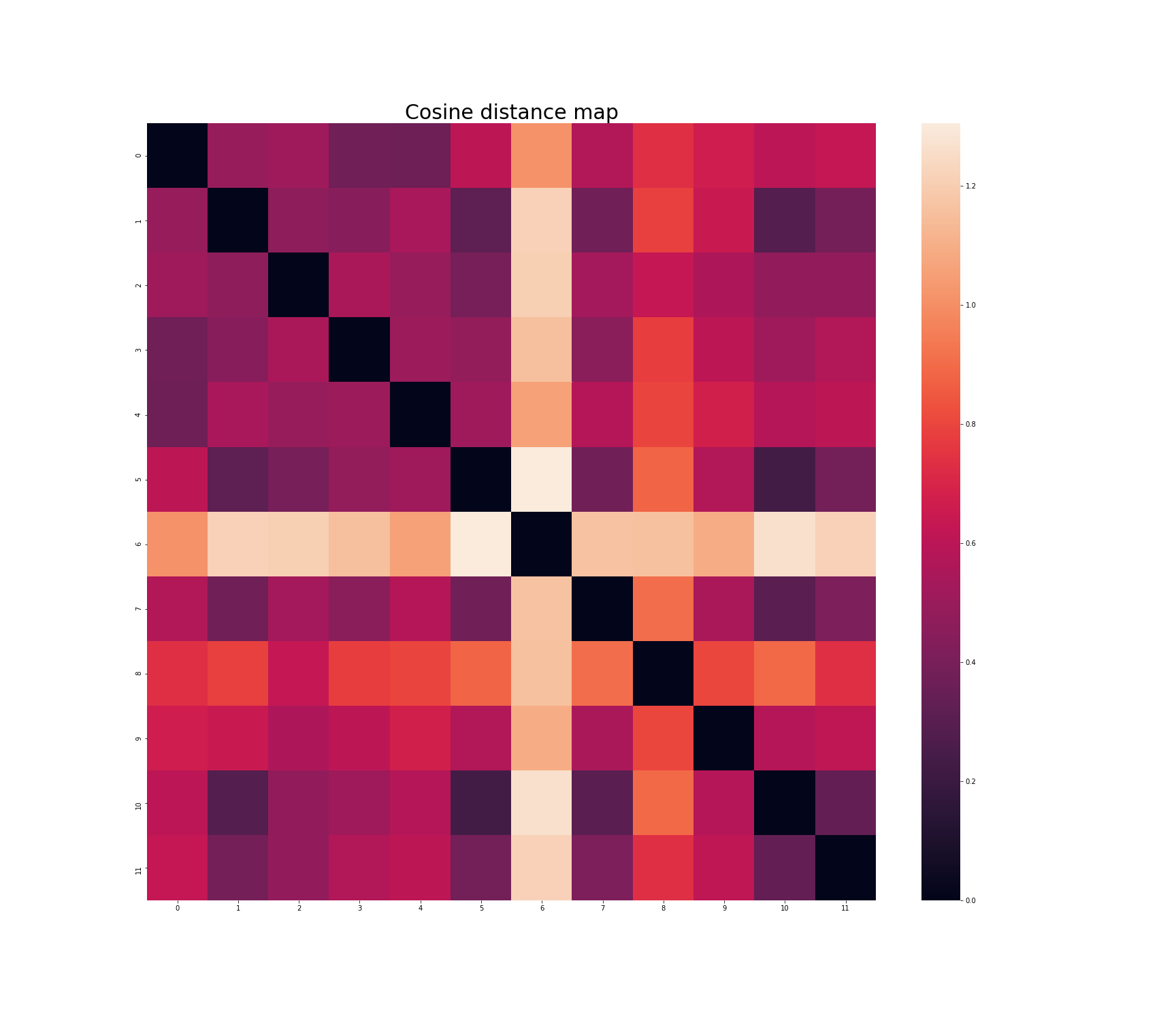}}
  \centerline{(b) Cosine Distance matrix $\mathcal{M}_{12\times12}$}\medskip
\end{minipage}

\caption{Ugly Ducklings and Cosine Distance. Fig (a): 12 lesions of a patient on the torso, along with target and outlier score; 1 indicates an ugly duckling. Fig (b): the cosine distance map of each lesion with other neighbouring/contextual lesions. The map shows the 7th and 9th lesions having high cosine distances, indicating that they are very dissimilar and hence they are labelled as ugly ducklings.}
\label{fig:fig1}

%
\end{figure*}

\section{Integration of IPCA in learning Models}
We propose a CAD system that incorporates IPCA (to identify ugly ducklings) and uses its results for deriving the final  diagnosis of melanoma. 
\subsection{Identifying Ugly Ducklings}

The task at hand is to identify the ugly ducklings (or outliers) among a set of contextual lesions (i.e. from  one region) for a patient. Let $R$ be the region of a patient with lesions $L_{i}$, ($i \in [1,N], i \in \mathbb{Z}$). Given a  set of lesions for $R$, for every $L_i$, the corresponding contextual lesions are nothing but $L_{j}, \forall j \in [1,N], j \neq i$. In order to determine if a lesion is an outlier or not, we propose to use a score based on the cosine distance between $L_i$ and the contextual lesion $L_j$ to characterise $L_i$. The details of the proposed methods are presented next.

Features of $L_i$ are extracted using a CNN pretrained on a melanoma dataset (The ISIC 2019 \cite{isic20191, isic20192, isic20193} was used in our work). Next, a global average pooling is applied on these feature maps to obtain a feature vector $\pmb{g_{i}}$.

 We use the cosine distance of a lesion from all contextual lesions to define a matrix  $\mathcal{M}(i,j);i,j \;\in [1,N]$ as 
\begin{equation}
\label{ref:distEq}
\begin{aligned}
    \mathcal{M}(i,j) &= 1 - \frac{\pmb{g_{i}} \cdot \pmb{g_{j}}}{\|\pmb{g_{i}}\|\times\|\pmb{g_{j}}\|}\\
\end{aligned}
\end{equation}

When a lesion, say $L_3$, is an outlier among $L_j; j\in[1,N]$ the choice of Cosine distance ensures that $\mathcal{M}(3,j)$ has a higher value than if it was similar to the other lesions in the set. 

 Given a set of contextual lesions, the average value of the cosine distances is a good way to characterise any lesion in the set. Thus, given a lesion $L_{i}$ its outlier score is defined as  
\begin{equation} \label{ref:outlierEq}
    \mathcal{O}_{i} = 
    \frac{\sum_{j=1, j\neq i}^{j=N} \mathcal{M}(i,j)}{N-1} 
\end{equation}

From the above definition, it is seen that $ \mathcal{O}_{i} \in [0,1]$; in practice, $\mathcal{O}_{i}=0$ is possible only when all the contextual lesions are duplicates. The robustness of any outlier detection will depend on the sample size $N$ as too small an $N$ can lead to misleading results. Hence, empirically, the required $N$ was taken to be 6.  When $N \leq 5$, IPCA is not possible, and we treat $L_i$ as an outlier with $ \mathcal{O}_{i} = 1$. This definition will ensure that when the system does not have access to a patient's other neighbouring lesions, every lesion should be suspected and decision should be based only on morphological analysis or LFA. 

Finally, the outliers are identified using an analysis based on the Inter Quartile Range ($IQR$) which is the difference between first ($Q_{1}$) and third quartiles ($Q_{3}$) of $\mathcal{O}_{i}$. A lesion $L_{i}$ is an outlier if,
\begin{equation} \label{ref:iqr_eq}
        \mathcal{O}_{i} \geq Q_{3} + k \times IQR 
\end{equation}
where, the value of ${k}$ determines the tolerance of the outlier detector. Higher the value of ${k}$, lower is the tolerance. Fig \ref{fig:fig1}. shows some sample contextual leasions and a visualisation of the $\mathcal{O}$ matrix as an image. 

\subsection{Proposed CAD system}
\begin{figure}[hb]
\begin{minipage}[b]{1.0\linewidth}
  \centering
  \centerline{\includegraphics[width=8.5cm]{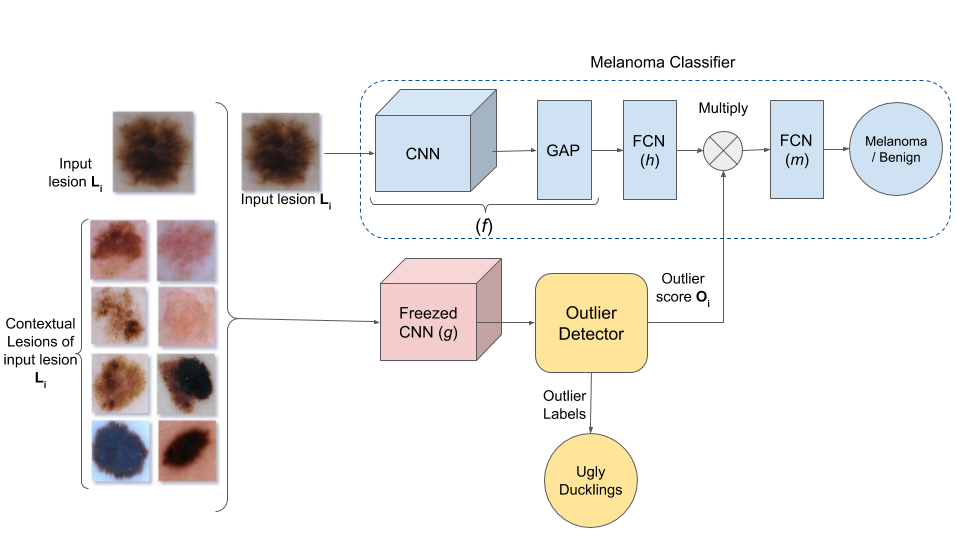}}
  \medskip
\end{minipage}
\caption{Proposed CAD system for melanoma detection}
\label{fig:networkarch}
\end{figure}

We now turn to the CAD system design. A schematic of the proposed design is shown in Fig. \ref{fig:networkarch}. The top pipeline is a melanoma classifier whose decision is to be moderated based on the outlier score. This is done as follows.

Assume a CNN based classifier $f$ with embeddings, $f(L_{i})$ derived after a Global Average Pooling. These are passed through a fully connected network (FCN), $h$ to obtain $h(L_{i})$. The outlier scores, $\mathcal{O}_{i}$ are used to weigh these features by performing $h'(L_{i}) = \mathcal{O}_{i} \cdot h(f(L_{i}))$, so that the ugly ducklings are learnt with more importance. In order to obtain the final decision, a binary classification is done based on $h'(L_{i})$ to predict labels, $p_{i}$ using a FCN $m$, such that $p_{i} = m(h'(L_{i}))$,  as shown in Fig. \ref{fig:networkarch}.

There are two ways to use the outlier scores, $\mathcal{O}_{i}$: (a) The score can be injected into the network to give an increased weight to ugly ducklings and penalise misclassification, or, (b) it can be directly included in the loss function. Method (a) ensures that the outlier scores is used for inference, whereas (b) guarantees its role only during the learning phase. We chose (a).

The objective of the system is to minimize loss ($\mathcal{L}$) between the ground truth distribution of lesion labels, $y_{i}$ and $p_{i}$ for lesion $L_{i}$ by finding optimal network weights $f_{w}^{*},h_{w}^{*},m_{w}^{*}$ for networks $f, h, m$ such that:
\begin{equation}\label{ref:objective}
    f_{w}^{*},h_{w}^{*},m_{w}^{*} = \argmin_{f_{w},h_{w},m_{w}}\;\mathcal{L}(y_{i},p_{i})
\end{equation}
Consider the networks, $f,h,m$. If the inputs are $x_{f},x_{h},x_{m}$ and network weights are $f_{w},h_{w},m_{w}$ respectively, then by chain rule,
\begin{equation}
\begin{aligned}
    \frac{\partial \mathcal{L}(y_{i},p_{i})}{\partial m_{w}} &= \frac{\partial \mathcal{L}(y_{i},p_{i})}{\partial p_{i}}\cdot\frac{\partial p_{i}}{\partial m_{w}}\\
    \frac{\partial \mathcal{L}(y_{i},p_{i})}{\partial h_{w}} &= \frac{\partial \mathcal{L}(y_{i},p_{i})}{\partial p_{i}}\cdot\frac{\partial p_{i}}{\partial x_{m}}\cdot\frac{\partial x_{m}}{\partial h_{w}}\\
    \frac{\partial \mathcal{L}(y_{i},p_{i})}{\partial f_{w}} &= \frac{\partial \mathcal{L}(y_{i},p_{i})}{\partial p_{i}}\cdot\frac{\partial p_{i}}{\partial x_{m}}\cdot\frac{\partial x_{m}}{\partial x_{h}}\cdot\frac{\partial x_{h}}{\partial f_{w}}\\
\end{aligned}
\end{equation}
Substituting $p_{i} = m(x_{m})$, $x_{m} = \mathcal{O}_{i}\cdot h(x_{h})$, $x_{h} = f(x_{f})$ and $x_{f} = L_{i}$, we have,
\begin{equation}\label{ref:grad_scale}
    \begin{aligned}
    \frac{\partial x_{m}}{\partial h_{w}} &= \mathcal{O}_{i}\cdot\frac{\partial h(x_{h})}{\partial h_{w}}\\
    \frac{\partial \mathcal{L}(y_{i},p_{i})}{\partial h_{w}} &= \mathcal{O}_{i}\cdot\frac{\partial \mathcal{L}(y_{i},p_{i})}{\partial p_{i}}\cdot\frac{\partial p_{i}}{\partial x_{m}}\cdot\frac{\partial h(x_{h})}{\partial h_{w}}\\
    \frac{\partial \mathcal{L}(y_{i},p_{i})}{\partial f_{w}} &= \mathcal{O}_{i}\cdot\frac{\partial \mathcal{L}(y_{i},p_{i})}{\partial p_{i}}\cdot\frac{\partial p_{i}}{\partial x_{m}}\cdot\frac{\partial h(x_{h})}{\partial x_{h}}\cdot\frac{\partial x_{h}}{\partial f_{w}}\\
    \end{aligned}
\end{equation}
Thus, from equation \ref{ref:grad_scale} we can see that $\mathcal{O}_{i}$ scales the gradients of the networks. The higher value of \ref{ref:grad_scale} for an outlier, therefore will ensure that the ugly ducklings are penalised more for misclassification, yielding the optimal network weights $f^{*}_{w},h^{*}_{w},m^{*}_{w}$ according to the objective function in equation \ref{ref:objective}.

A key strength of the proposed design is that  any existing CAD system for melanoma detection can be plugged in place of the melanoma classifier.

\section{Experiments}
\label{sec:experiments}
\subsection{Dataset}
Experiments were done using the ISIC 2020 Challenge Dataset \cite{isic2020}. This dataset provides patient-level contextual information and has 33,126 dermoscopic images of 2056 patients from Europe, North America and Australia with an average of 16 lesions per patient and 584 confirmed melanomas.

\subsection{Dataset Preprocessing and Augmentation}

All lesion images were resized to a fixed size ($512 \times 512 \times 3$) and normalized using ImageNet statistics. Augmentation was applied via affine transformations  (random shifts, shear, zoom, rotation, horizontal and vertical flips), random change of saturation, contrast, and brightness. This was done on the train, test, and validation sets. In all our experiments, test time augmentations with 22 steps was done on both test and validation sets during inference.

\subsection{Implementation Details}

\textbf{Computing outlier scores:} In ISIC 2020 dataset, not every patient has multiple images/lesions that are sufficient enough to find ugly ducklings. Hence, only patients who have more than 5 contextual lesion images were considered for outlier detection assessment. The features of each of these lesions belonging to a patient in a given location was extracted by using an EfficientNet B6 network \cite{effnet} pretrained on the ISIC 2019 dataset. These features were then used to compute outlier scores as given by equation \ref{ref:outlierEq}. Empirical results showed that the outlier detector performed better when $k$ in equation \ref{ref:outlierEq} is set to 1.


Different variants of the  proposed system in Fig. \ref{fig:networkarch} was created by changing the classifier module : EfficientNet B6 Noisy-Student (NS) \cite{noisy_student}, EfficientNet B7 NS . All networks were trained  on TPU (Tensor Processing Units) with 8 TPU v3 cores and 128 GiB memory. In the interest of computation time, the  outlier/ugly duckling labels were precomputed to serve as an additional information to CNN classifier as shown in Fig \ref{fig:networkarch}. A group 5-fold split was applied on patient ID to ensure that train and validation sets have no patient in common. This prevented data leakage.

The Rectified Adam optimizer \cite{radam} was used with a Focal-Loss based cross entropy loss \cite{focal_loss} function with an initial learning rate as 3$e$-3 which is reduced on plateau. Each fold was trained for 25 epochs and early stopping was used to prevent any overfitting. 
\begin{table}[h!]
\centering
\begin{tabular}{ |c|c|c|c|c|}
 \hline
 {\Longunderstack{Model}}&
\multicolumn{2}{|c|}{\Longunderstack{Without\\ Ugly Ducklings}} &
 \multicolumn{2}{|c|}{\Longunderstack{With\\ Ugly Ducklings}} \\\cline{2-5}
 
 &Sen. &Spec. &Sen. &Spec.\\
 \hline
{\Longunderstack{EfficientNet\\-B6 Noisy Student}}	&0.89	&0.82 &\textbf{0.91} &\textbf{0.84}\\
 \hline
{\Longunderstack{EfficientNet\\-B7 Noisy Student}}	&0.84	&0.81 &\textbf{0.89} &\textbf{0.85}\\
  \hline
 {\Longunderstack{ResNet 101}}	&\textbf{0.91}	&0.74 &0.88 &\textbf{0.82}\\
  \hline
  
   {\Longunderstack{VGG 19}}	&0.87	&0.72 &\textbf{0.87} &\textbf{0.77}\\
  \hline
  
 \end{tabular}
 \caption{Sensitivity \& Specificity of models without and with Ugly ducklings information.}
 \label{ref:effRes}
 \end{table}

\subsection{Results and Discussion}
First we report on the outlier detection performance. When the cosine distance based scores are treated as class scores, the attained performance was 0.82 sensitivity and 0.81 specificity which is comparable to benchmarked deep learning models.

The sensitivity and specificity values of the proposed CAD system variants (see section \ref{sec:experiments}) were calculated at the \textit{knee point} of the ROC curve and are presented in Table \ref{ref:effRes}. The inclusion of ugly duckling information is seen to lead to a consistent boost in specificity and sensitivity. While the boost is by moderate amounts for variants with strong classifier models like EfficientNets, it is more significant for variants with weaker networks like VGG \cite{VGG}, ResNet \cite{resnet}. 
A 3\% decrease in sensitivity is more than offset by a boost of 10.8 \% in specificity. 

Overall, our experimental results support the emphasis of the clinical studies on the importance of the use of ugly ducklings in early detection of melanoma. While the boost in performance depends on the type of classifier used, the key advantage of our design is that the output of the system is a prediction label \textit{together with} an outlier score which, in the case of contextual lesions, serves as a much needed confidence measure to clinicians. 

The current benchmark on ISIC 2020 dataset is 0.949 Area under the ROC (AUC) score, which is recorded in a Kaggle competition, launched jointly by SIIM (Society for Imaging Informatics in Medicine) and ISIC (International Skin Imaging Colloboration). According to \cite{siim_org_2020}, the top scoring team ensembled multiple EfficientNet models (B3-B7), SE-ResNeXt-101 and ResNeXt-101 with various input resolutions ranging from 384$ \times $384 to 1024$ \times $1024 along with meta data. Only patient-level information, in the form meta-data, was used for the diagnosis of melanoma, rather than the patient's contextual images. In contrast, our approach, when implemented with an simple ensemble of weighted average of scores of only two EfficientNet B6 and B7 Noisy student models on 512x512 images, achieved an AUC  of \textbf{0.940} and \textbf{0.935} for inference with and without integration of ugly ducklings information, respectively. These results are promising and indicate the potential for improvements with more extensive ensembling of various models.





\section{Conclusion}
\label{sec:conclusion}
We presented an approach based on the use of contextual information in clinical practice to assess melanoma. Specifically we are identifying outliers and using them jointly with visual features to assess melanoma classification. 

The key strength of the proposed approach is that it provides a clinically meaningful explanation for a decision by including an outlier score. The proposed design permits a flexible integration of outlier detector with any existing architecture. It needs to be emphasised that the design includes both an outlier detector and a classifier to flexibly handle a range of scenarios and cases: cases where information about contextual lesions is available or not as well as cases with isolated lesions.

The experimental results indicate that the inclusion of Ugly duckling method in CAD systems is beneficial both in terms of detection performance boost and in providing a supportive clinical explanation when apt.



\bibliographystyle{IEEEbib}
\bibliography{refs,strings}

\end{document}